\documentclass[pdflatex,sn-mathphys-num]{sn-jnl}

\usepackage{graphicx}%
\usepackage{multirow}%
\usepackage{amsmath,amssymb,amsfonts}%
\usepackage{amsthm}%
\usepackage{mathrsfs}%
\usepackage[title]{appendix}%
\usepackage{xcolor}%
\usepackage{textcomp}%
\usepackage{manyfoot}%
\usepackage{booktabs}%
\usepackage{algorithm}%
\usepackage{algorithmicx}%
\usepackage{algpseudocode}%
\usepackage{listings}%
\usepackage{enumitem}
\setlist[description]{style=nextline}
\usepackage{amsmath}

\theoremstyle{thmstyleone}%
\theoremstyle{thmstyletwo}%

\theoremstyle{thmstylethree}%

\begin{document}

\title[Article Title]{"Pass the butter": A study on desktop-classic multitasking robotic arm based on advanced YOLOv7 and BERT}


\author[1]{\fnm{Haohua} \sur{Que}}\email{qh13005968844@gmail.com}
\equalcont{These authors contributed equally to this work.}

\author[2]{\fnm{Wenbin} \sur{Pan}}\email{wenpin@bjfu.edu.cn}
\equalcont{These authors contributed equally to this work.}

\author[1]{\fnm{Jie} \sur{Xu}}\email{xu\_jie@bjfu.edu.cn}

\author[3]{\fnm{Hao} \sur{Luo}}\email{miku@mails.cust.edu.cn}

\author*[1]{\fnm{Pei} \sur{Wang}}\email{wangpei@bjfu.edu.cn}

\author*[1]{\fnm{Li} \sur{Zhang}}\email{zhang\_li@bjfu.edu.cn}

\affil*[1]{\orgdiv{School of Science}, \orgname{Beijing Forestry University}, \orgaddress{\street{35 Qinghua E Rd}, \city{Beijing}, \postcode{100083}, \state{Beijing}, \country{China}}}

\affil[2]{\orgdiv{School of Humanities and Social Science}, \orgname{Beijing Forestry University}, \orgaddress{\street{35 Qinghua E Rd}, \city{Beijing}, \postcode{100083}, \state{Beijing}, \country{China}}}

\affil[3]{\orgdiv{School of Computer Scinece and Technology}, \orgname{Changchun University of Science and Technology}, \orgaddress{\street{No. 7186, Weixing Road}, \city{Changchun}, \postcode{130012}, \state{JiLin}, \country{China}}}


\abstract{In recent years, various intelligent autonomous robots have begun to appear in daily life and production. Desktop-level robots are characterized by their flexible deployment, rapid response, and suitability for light workload environments. In order to meet the current societal demand for service robot technology, this study proposes using a miniaturized desktop-level robot (by ROS) as a carrier, locally deploying a natural language model (NLP-BERT), and integrating visual recognition (CV-YOLO) and speech recognition technology (ASR-Whisper) as inputs to achieve autonomous decision-making and rational action by the desktop robot.
Three comprehensive experiments were designed to validate the robotic arm, and the results demonstrate excellent performance using this approach across all three experiments. In Task 1, the execution rates for speech recognition and action performance were 92.6\% and 84.3\%, respectively. In Task 2, the highest execution rates under the given conditions reached 92.1\% and 84.6\%, while in Task 3, the highest execution rates were 95.2\% and 80.8\%, respectively. Therefore, it can be concluded that the proposed solution integrating ASR, NLP, and other technologies on edge devices is feasible and provides a technical and engineering foundation for realizing multimodal desktop-level robots.}




\maketitle

\section{Introduction}\label{sec1}

\begin{figure}[h]
\includegraphics[width=\textwidth]{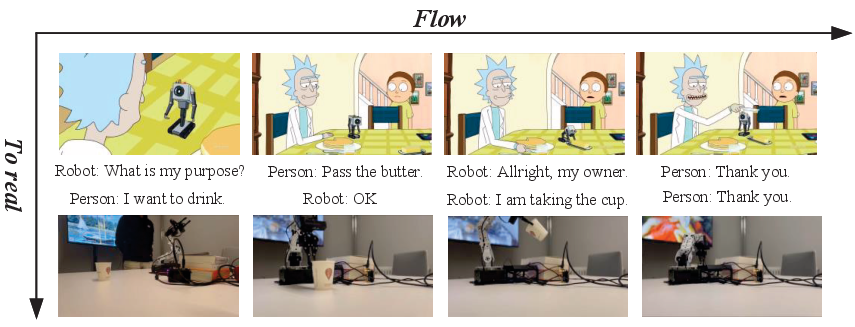}
\caption{The flow of robot execution motion (Animation and Reality).}\label{fig:The flow of robot execution motion (Animation and Reality)}
\end{figure}

Historically, natural language models were restricted to text and did not account for the physical world, often providing impractical responses to tasks they could not physically execute, such as passing butter. Some science fiction films and series, such as Rick and Morty, have already featured similar applications, such as the butter-passing robot, and it's an example relevant to this paper's work which shows in Fig. \ref{fig:The flow of robot execution motion (Animation and Reality)} Incorporating visual inputs allows robots to analyze their environments and make spatial decisions, effectively transitioning AI from theoretical applications to practical implementations\cite{1}. This integration necessitates considering additional real-world elements in their planning, improving the feasibility of their responses\cite{2}.

\begin{figure}[h]
\includegraphics[width=\textwidth]{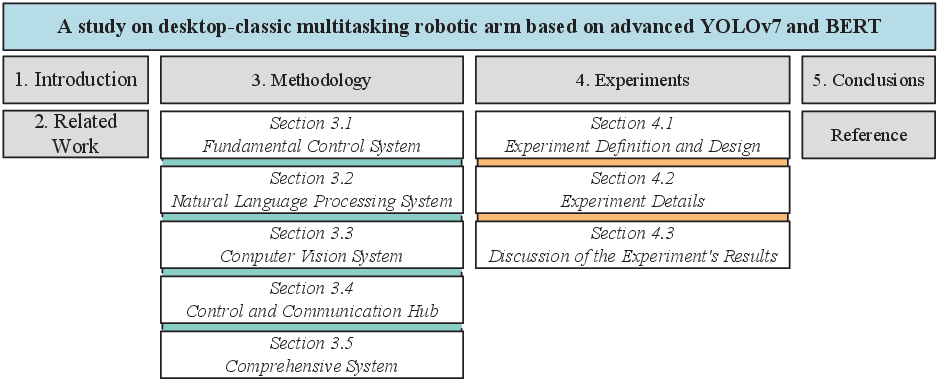}
\caption{The structure of paper.}\label{fig:The structure of paper}
\end{figure}

Dynamic planning enabled by natural language models is crucial for safe and flexible robot operations. It involves making optimal decisions for moving and manipulating objects in varying environments, adjusting paths in real-time to navigate around obstacles and ensuring task completion. This capability tests both the robot's real-time processing and its ability to predict and adapt to environmental changes\cite{A3}.
Handling multimodal inputs presents significant computational challenges, particularly for lightweight, desktop devices that require precise resource allocation. In practice, robots perform both real-time tasks like locating objects and controlling movements, and less urgent, non-real-time tasks. The latter can be offloaded to more powerful computing resources such as central processing units and graphics cards in standard personal computers. For instance, high-demand tasks such as processing user commands and language analysis are offloaded to a remote computer equipped with a graphics card, which handles intensive computations like language intent analysis through models like Whisper and BERT\cite{RecentAdvancesinNaturalLanguageProcessing}.
At the remote computing node, data from robot-mounted sensors is processed over a network, converting audio to text via the Whisper model and recognizing intents with BERT, thus controlling the robot based on natural language commands\cite{Chen2023}.
This paper proposes simplifying complexity by assigning computationally intensive and non-critical tasks to remote systems, while maintaining essential control tasks on the robot itself. This strategy allows desktop-level robots to comprehend natural language and autonomously execute actions, details of which are elaborated in the structure depicted in Fig. \ref{fig:The structure of paper}.

\section{Related Work}\label{sec2}
As previously mentioned, to facilitate the operation and use of desktop-level robots with multimodal inputs, the engineering aspects of this paper can be divided into speech recognition, natural language processing, and robot control components. Below is an analysis of the related work for each component.

\subsection{ASR Model}
The ASR model, such as the end-to-end ASR model Whisper, plays a crucial role in voice recognition. A study on the effectiveness of contextual biasing when using this model introduced a specific tree-constrained pointer generator (TCPGen) component and a dedicated training scheme to dynamically adjust the final output without modifying the Whisper model parameters\cite{zheng2021adapting}. Experiments across three datasets showed a significant reduction in errors for bias words when using a list of 1000 bias words\cite{arumugam2019grounding}. Especially in domain-specific data, contextual biasing was more effective, improving the performance of Whisper and GPT-2 without sacrificing their generality \cite{End-to-EndSpeechRecognition}. The conclusion emphasized contextual biasing as an effective means to enhance the recognition performance of domain-specific vocabulary while maintaining the model's generalization ability\cite{Whisper}.

\subsection{BERT Model}
BERT (Bidirectional Encoder Representations from Transformers) is a pre-trained deep learning model introduced by Google in 2018\cite{BERT,HumanLanguageUnderstanding}. It has garnered widespread attention in the NLP field due to its state-of-the-art results on various language understanding tasks at the time\cite{BERT}. One of the core innovations of BERT is the use of a bidirectional encoder representation from the Transformer model to more comprehensively understand the context of language.

\subsubsection{principles of the BERT model}
\textbf{1. Bidirectional Transformer Encoder:} BERT employs the encoder architecture of the Transformer, a model based on self-attention mechanisms that capture global dependencies in input data. Unlike previous models, BERT considers the context from both sides of a word, thereby obtaining a deep bidirectional representation of the word\cite{MovshovitzAttias2013NaturalLM}.

\textbf{2. Pre-training and Fine-tuning:} BERT’s training is divided into two stages. It undergoes pre-training on a large text corpus to learn a universal representation of language, utilizing tasks such as Masked Language Model (MLM) and Next Sentence Prediction (NSP)\cite{shi-demberg-2019-next}. It is then fine-tuned on specific downstream tasks, like question answering, text classification, and named entity recognition, by adjusting the pre-trained model with a small amount of task-specific data.

\textbf{3. Masked Language Model (MLM):} During pre-training, some words in the input sequence are randomly replaced with a special [MASK] token, and the model predicts these masked words based on context\cite{Salazar_2020}.

\textbf{4. Next Sentence Prediction (NSP):} BERT learns sentence-level relationships by predicting whether two sentences are adjacent in the original text, improving the model's ability to handle tasks that require understanding the relationship between sentences, such as paragraph or dialogue processing\cite{LanguageBind}.

\section{Methodology}\label{sec3}

\subsection{Fundamental Control System}\label{subsec4}
To achieve quantitative control of a robotic arm, it is necessary to model the arm. The kinematics of a 6-degree-of-freedom (6dof) robotic arm can be described using the Denavit-Hartenberg (DH) parametrization method\cite{hao20116}. This approach employs a set of parameters to define the geometric characteristics of the robotic arm and the relationships between changes in its joints. The following provides a more detailed display of the kinematic model of a six-axis robotic arm\cite{hasan2006adaptive}.

Coordinate System Definition: The base coordinate system of the robotic arm and the coordinate system for each joint are defined. Typically, the base coordinate system, $O_0$, coincides with the world coordinate system, and each joint's coordinate system, $O_i$(i = 1,2,...,6), is located on the joint axis, with the $z_i$-axis aligned with the joint's rotation axis\cite{ROS2}.
DH Parameters: The DH parameters are used to describe the geometric relationships between adjacent coordinate systems. For a six-axis robotic arm, the DH parameters are as follows:
\\
\\
${\alpha_i}$: The rotation angle of the ${z_{i-1}}$-axis around the ${x_i}$-axis.

${a_i}$: The distance from the ${z_{i-1}}$-axis to the ${z_i}$-axis, projected along the ${x_i}$-axis.

$d_i$: The distance from the $x_{i-1}$-axis to the ${x_i}$-axis, projected along the ${z_{i-1}}$-axis.

$\text{$\theta$}_i$: The rotation angle of joint $i$, around the ${z_{i-1}}$-axis.
\\

Transformation Matrix \eqref{eq1}: For each joint, the transformation matrix relative to the previous coordinate system can be calculated based on the DH parameters. For example, the transformation matrix for the 
$i$ joint is:

\leavevmode \\
\begin{equation}
\label{eq1}
A_{i}=\left[\begin{array}{cccc}
\cos \left(\theta_{i}\right) & -\sin \left(\theta_{i}\right) & 0 & a_{i} \\
\sin \left(\theta_{i}\right) \cos \left(\alpha_{i}\right) & \cos \left(\theta_{i}\right) \cos \left(\alpha_{i}\right) & -\sin \left(\alpha_{i}\right) & -\sin \left(\alpha_{i}\right) d_{i} \\
\sin \left(\theta_{i}\right) \sin \left(\alpha_{i}\right) & \cos \left(\theta_{i}\right) \sin \left(\alpha_{i}\right) & \cos \left(\alpha_{i}\right) & \cos \left(\alpha_{i}\right) d_{i} \\
0 & 0 & 0 & 1
\end{array}\right]
\end{equation}
\\
End-Effector Position and Orientation: By multiplying the transformation matrices of all joints, the transformation matrix of the end-effector relative to the base coordinate system, denoted as $T_{06}$, can be obtained. Information regarding the position and orientation of the end-effector can be extracted from this matrix. Table \ref{tab1} is the value of the DH parameter corresponding to each joint.

\begin{table}[h]
\caption{DH parameter}\label{tab1}%
\begin{tabular}{@{}llllll@{}}
\toprule
$i$ & ${\alpha_{i-1}}$  & ${a_{i-1}}$   & ${d_i}$ & ${{\theta}_i}$    & ${\theta}_{range}$ \\
\midrule
row 1    & 0   & 0  & 0  & $\theta_1$   & $\theta_1(-120, 120)$ \\
row 2    & -90   & 0  & 0  & $\theta_2$  & $\theta_2(-180, 0)$ \\
row 3    & 0   & 0.12941763737  & 0 & $\theta_3$  & $\theta_3(-120, 120)$ \\
row 4    & 0   & 0.12941763737  & 0 & $\theta_4$  & $\theta_4(-200,20)$ \\
row 5    & -90   & 0  & 0   & $\theta_5$  & $\theta_5(-120, 120)$\\
\botrule
\end{tabular}
\end{table}

\begin{figure}[H]
\includegraphics[width=\textwidth]{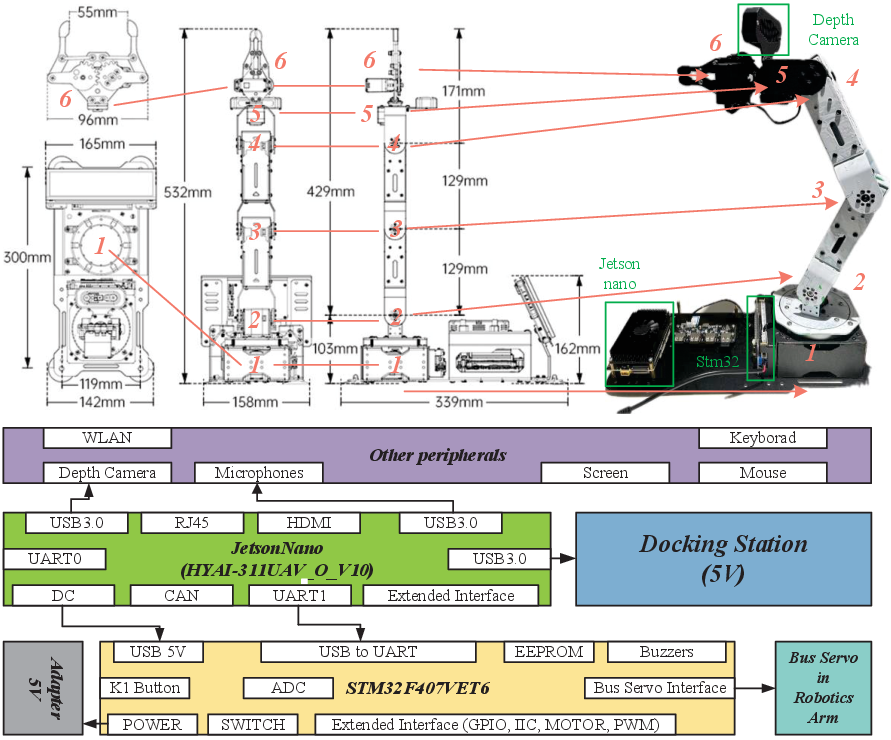}
\caption{Robotic arm structure diagram and device tree.}\label{fig:Robotic arm structure diagram and device tree}
\end{figure}

Fig. \ref{fig:Robotic arm structure diagram and device tree} illustrates the structural diagram and the device tree of the robotic arm. Utilizing the STM32F07VET6, the kinematic modeling of the 6dof robotic arm was accomplished, with communication established via serial port and with the Jetson\cite{iqbal2012modeling}. Bi-directional communication between the Jetson and the STM32 allows for the execution of specific tasks: upon receiving nodal coordinate commands from the Jetson, the STM32 calculates the angles for each associated nodal servo through inverse kinematics and sends this information back to the Jetson; conversely, when given commands for the angles of each node by the Jetson, the STM32 employs forward kinematics to determine the coordinates of each node and relays this information back to the Jetson\cite{Chen2023}. The commands issued by the Jetson can selectively instruct the STM32 to either perform actual movements or merely conduct kinematic information queries\cite{hasan2006adaptive}.

\subsection{Natural Language Processing System}
Natural Language Processing (NLP) systems transform natural language from text and audio into commands that robots can understand and execute, significantly improving robots' ability to interact naturally and intelligently with humans\cite{RecentAdvancesinNaturalLanguageProcessing}. For text inputs, the NLP system analyzes the text, denoted as $T(n) = \{W1,W2,W3,...,Wn\}$, to identify specific intents\cite{AttentionisAllyouNeed}. There are $K$ kinds of intents, represented as $X_i$ ($i=1,2,3,...,k$), each with a probability $p(X_i)=F( T(n) )$. The function $F(x)$ utilizes the BERT pre-training model to obtain the semantic vector of the sentence, which is then input into a Multi-Layer Perceptron (MLP) to perform intent classification: $F(T(n)) = MLP( BERT(T(n)))$. The intent with the highest probability is chosen: $p(X_m) = \max\{p(X_i)\}$. If $p(X_m)$ exceeds a certain threshold, the intent of the sentence is determined to be $X_m$\cite{A3}.

For voice inputs, the audio, denoted as $Audio(m) =\{l1,l2,l3,...,lm\}$, is first transcribed into text using the Whisper model: $T(n) = Whisper(Audio(m))$. The transcribed text undergoes the same intent recognition and classification process as described above, facilitating the appropriate response from the robot\cite{BERT}.

The BERT (Bidirectional Encoder Representations from Transformers) model is pivotal in this process. It utilizes Transformer encoders to deeply understand language through a two-stage approach: pre-training and fine-tuning. Pre-training on extensive datasets allows BERT to grasp diverse linguistic nuances and relationships\cite{4}. The model processes texts bidirectionally during this stage, enhancing its contextual understanding\cite{HumanLanguageUnderstanding}.

In the fine-tuning stage, BERT is tailored to specific NLP tasks, such as text classification or named entity recognition, based on task-specific requirements. This customization enhances BERT's effectiveness for particular applications.

\begin{figure}[H]
\includegraphics[width=\textwidth]{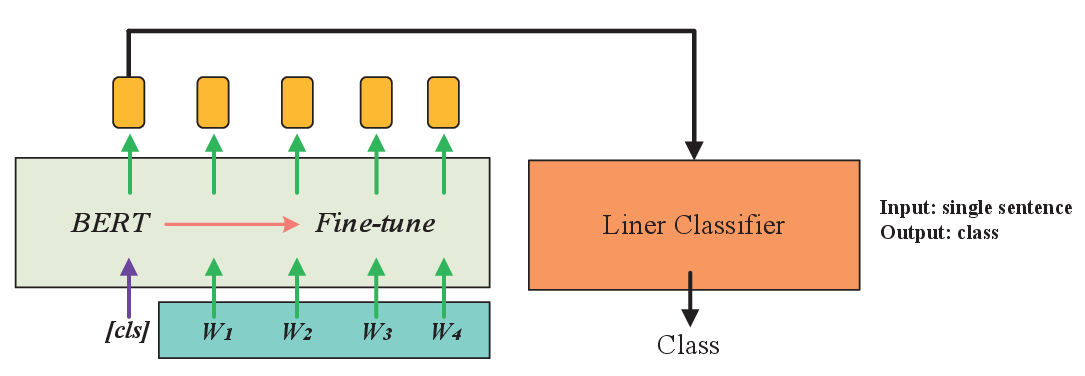}
\caption{BERT model fine-tuning for intent classification.}\label{fig:BERT model fine-tuning for intent classification}
\end{figure}

Google offers pre-trained BERT models in multiple languages, facilitating their fine-tuning by developers to expedite development and deployment. In this paper, we fine-tune the bert-base model for intent classification and slot filling. The process begins by adding a "[CLS]" token to the start of tokenized text, creating an extended vector that feeds into the BERT model. The resulting text features are then classified using a Multi-Layer Perceptron (MLP) network for intent detection, as detailed in Fig. \ref{fig:BERT model fine-tuning for intent classification}.

\begin{figure}
    \centering
    \includegraphics{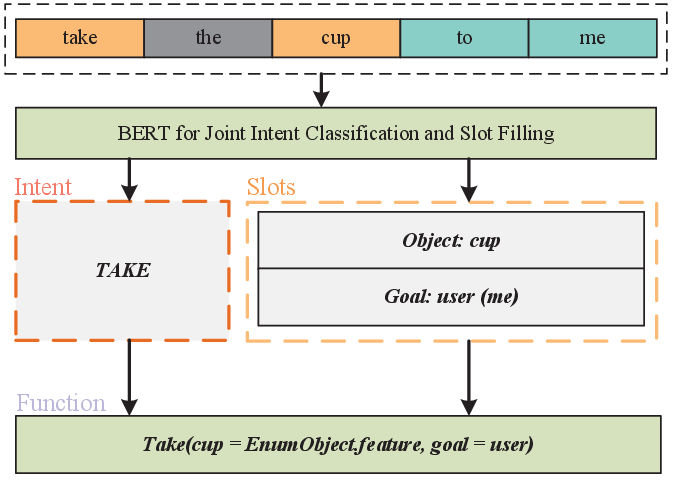}
    \caption{CMLP classification network.}
    \label{fig:MLP classification network}
\end{figure}

After inputting the text string into the intent classification and slot filling model, we obtain Intent and Slots, where Intent is the overall intent string of the text, and Slots are characters of intent-related information in the text. Finally, using a string matching method to functionalize the text's Intent string and transform the Slots strings into parameters for a computer-programmable function\cite{6}. The process is illustrated in the Fig\ref{fig:MLP classification network}.

\subsection{Computer Vision System}
Computer Vision (CV) is a critical technology in robotics, using cameras and sensors to gather image or video data for analysis, allowing robots to perceive and understand their environments \cite{yolo}. This incorporation of computer vision into robotics is essential for accurate object detection and manipulation. The YOLO (You Only Look Once) algorithm is particularly effective for such tasks, offering high-speed and precise object detection by treating detection as a regression problem\cite{Jian2010StatusAD}. This method directly predicts object bounding boxes, categories, and confidence scores from the image data\cite{2020IEEE,2022IEEE}.

In the specific mathematical implementation, for an input RGB image $I$, there are $K$ detectable categories of objects in the image, where $X_i$ ($i=1,2,3,...,k$) represents the $i$th type of object. The YOLO model is used to detect bounding boxes, categories, and confidence scores in the image, represented as ${ (x,y,w,h), X_i, p(X_i)}=YOLO(I)$. The center point of the bounding box is calculated as $(X_c,Y_c) = (x+\frac{1}{2}w, y+\frac{1}{2}h)$. This completes object detection and positioning on the two-dimensional image, obtaining the position $(X_c,Y_c)$ of the object $X_i$. Combined with the depth information $Z_c$ obtained from the depth camera at this point, the three-dimensional coordinate vector $T=(X_c,Y_c,Z_c)$ is obtained.

The forward kinematics matrix of the robotic arm is denoted as $M=A_1A_2...A_nM_c$, where $A$ is the forward kinematics matrix of the robotic arm, and $M_c$ is the spatial transformation matrix of the robotic arm end moving to the camera position. Assuming the camera to be orthogonal, the homogeneous vector $T_1=(X_c,Y_c,Z_c,1)$ is constructed. The position of the object in space is then obtained as $T_2 = M·T_1$.

YOLO uses a fully convolutional network to capture multi-scale, semantically rich features, essential for detecting objects of varying sizes, as described in \cite{AStudyonGenerativeModelsforVisualRecognition}. Despite its proficiency in real-time detection, YOLO only provides two-dimensional object information. For spatial tasks like robotic manipulation, three-dimensional data is crucial. This gap is bridged by integrating depth cameras, which provide the necessary depth values for each pixel by measuring the distance to objects using reflected light \cite{wang2022yolov7}. This paper proposes an alignment method between depth and RGB images to compute precise three-dimensional coordinates, enhancing robotic interaction capabilities with their environment \cite{6}.

\begin{figure}[h]
\includegraphics[width=\textwidth]{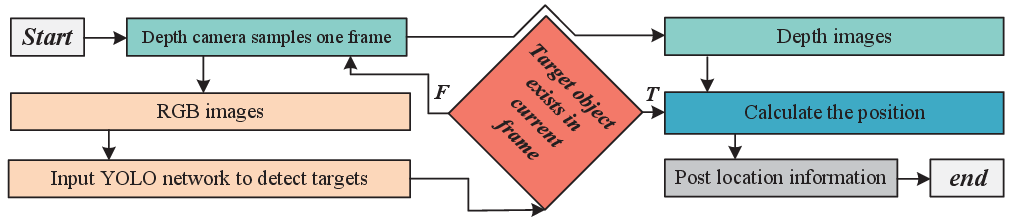}
\caption{The flow chart of the robot to achieve accurate target detection and grasp positioning.}\label{fig:The flow chart of the robot to achieve accurate target detection and grasp positioning}
\end{figure}

This integration allows for precise object detection and positioning, crucial for tasks such as robotic grasping. The complete implementation flow of this integrated system is detailed in Fig.\ref{fig:The flow chart of the robot to achieve accurate target detection and grasp positioning}, illustrating how robots can achieve enhanced target detection and manipulation in a three-dimensional space \cite{5}.

After adopting the method described above, which combines depth cameras for three-dimensional positional information of objects, stringent demands are placed on the computational power of edge devices. To enhance performance on embedded devices, we employed pruning and quantization techniques to reduce computational and storage overhead.

\begin{figure}[H]
\includegraphics[width=\textwidth]{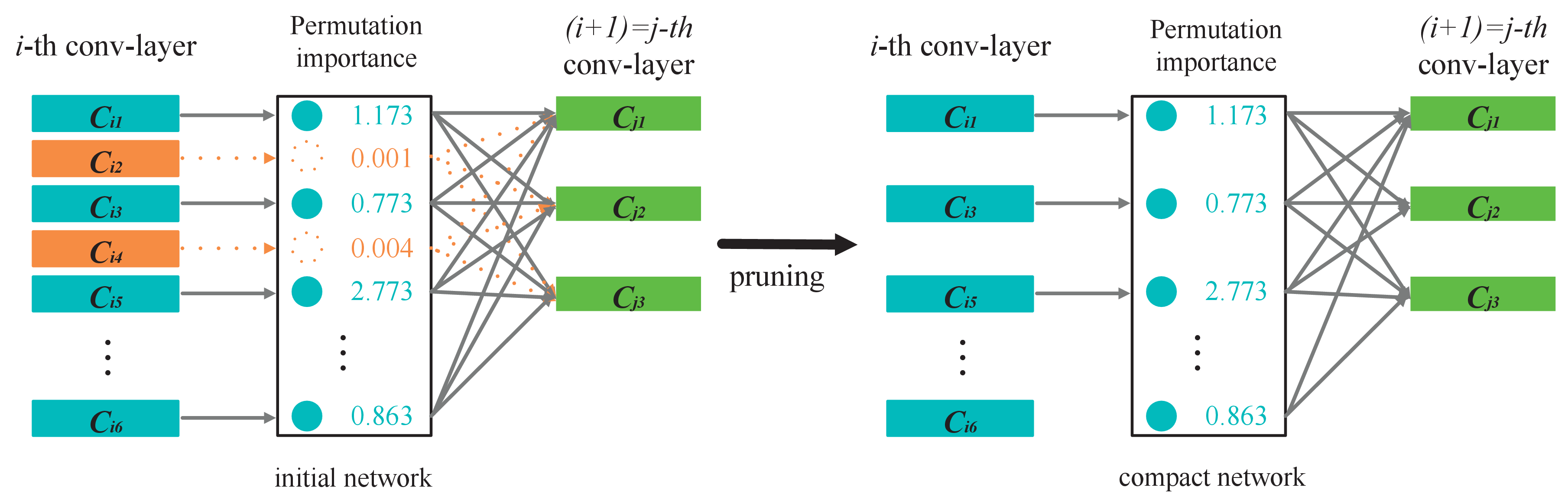}
\caption{The model underwent quantification of the pruning process.}\label{fig:The model underwent quantification of the pruning process}
\end{figure}

Pruning is a technique that reduces model size and computational demand by removing unnecessary parts of a neural network\cite{lin2020dynamic}.This work utilized a pruning method based on activation distribution and feature importance analysis, which shows in the Fig. \ref{fig:The model underwent quantification of the pruning process}. Initially, the importance of each neuron was determined by analyzing its activation distribution during forward propagation on the training set. Based on the significance of activation distributions, neurons with minimal impact on model performance were removed. Permutation Importance is introduced here as a metric to evaluate the significance of pruning and quantization, aiming to assess the importance of the convolutional layer in question. The importance of this convolutional layer is positively correlated with the value of Permutation Importance, with the calculation method for this value described by the following formula\eqref{eq3}.

\begin{equation}
\label{eq3}
    i_{j}=s-1 / K \sum_{k=1}^{K} s_{k, j}
\end{equation}

Permutation Importance is applicable to tabular data, where the assessment of feature importance is based on the degree to which the model's performance score decreases following the random permutation of the feature. Its mathematical expression can be represented as follows and here are the elucidations:

Input: A trained model $m$, and a dataset $D$ which could be a training, validation, or test set.

Model Performance Score $s$: The performance score of model $m$ on dataset $D$.
For each feature $j$ in dataset $D$, it indicates:

For each iteration $k$ in $K$ repetitions of the experiment, randomly permute feature $j$ to create a perturbed dataset $Dc_{k,j}$.
 
 Calculate the performance score $s_{k,j}$ of model $m$ on the perturbed dataset $Dc_{k,j}$.

The importance score of feature $j$, denoted as $i_j$, can then be represented by the formula\eqref{eq3} that calculates the average decrease in performance score across all $k$ repetitions for feature $j$.

During the pruning process, we also utilized feature   importance to analyze the weights of convolutional layers. By assessing the contribution of neurons to model accuracy and feature extraction capability, weights with minor impact on the model were identified and removed, further reducing model size and computational demand.

Additionally, we employed quantization to reduce the model's storage requirements and computational complexity. Quantization represents weights and activation values with fewer bits, reducing the storage space for model parameters\cite{zhou2023dataset}. This paper adopted fixed-point representation, converting the floating-point model's weights and activation values to fixed-point representation. This approach reduces computational demand without significantly lowering model performance.

After optimizing through pruning and quantization, the optimized model was deployed on embedded devices\cite{zhao2023atom}. Experiments evaluated the optimized model's performance metrics on embedded devices before and after optimization. Results indicated that by reducing model size and computational demand, this method achieved higher inference speed and efficiency on embedded devices while maintaining high detection accuracy.

\subsection{Control and Communication Hub}
After the robot is equipped with a basic control system, a computer vision system for visual object localization, and a natural language understanding system for comprehending human text and voice, it possesses the foundation required to perform complex actions\cite{ALightweightVisualSimultaneousLocalizationandMappingMethod}. Here have established a Control Communication Hub that enables the robot to complete complex tasks and communicate duplex with a remote host.

To facilitate the robot's execution of complex actions, we employ an action state machine. An action state machine is a formal model used to describe the behavior of a robot system. It defines and interprets the basic and special states within the system, detailing the transitions between states and the actions executed.Basic states represent the robot's behavior under normal operations, while special states describe the robot's behavior in exceptional or specific circumstances\cite{wang2022adaptive}. State transition diagrams are utilized to visualize the states and their transition relationships, aiding in understanding and designing the robot system\cite{ROS2}.

In the robot's action state machine, basic states consist of foundational actions and durations. For our mechanical arm, basic states might include actions such as "idle," "search," and "grab." Each basic state is associated with specific actions performed by the mechanical arm; for example, during the "search" state, the arm carries out the task of locating the target, while in the "idle" state, the arm waits for commands from the control host.

Special states are conditions in which the robot operates under exceptional or specific situations. Actions in these states include "arbitrary," "fault," "stuck," or "collision." The definition and handling of special states depend on the requirements of the task, the specific design of the mechanical arm, and its application. For instance, transitions from any state to a target state are conditionally evaluated for each "arbitrary" state at the end of every action, entering state transition; when a fault is detected, the arm enters the "fault" state and sends an alert to the control host\cite{Kapelyukh_2023}.

Extended states form a network of basic states and transition conditions, capable of executing complex commands, such as "follow target and grab." Extended states are realized by defining a series of basic states and state transitions to implement complex behaviors\cite{Artificialcognitionforsocialhuman–robotinteraction}. This combination allows the robot system to possess advanced functions and flexibility.

State transitions describe the relationships between different states of the robot. When the robot receives a task from the "idle" state, it transitions to the "grab" state. State transitions are usually triggered by specific conditions, often activated by sensor detection of specific events or external commands. Each state transition is accompanied by actions; for example, when the arm is detected to be stuck, the transition to the "stuck" state condition is met, entering the "stuck" state and notifying the control host for resolution.Global variables store dependency information for some states, such as the "action: move to target," which relies on a target coordinate variable in basic states\cite{2}.

\begin{figure}[H]
\includegraphics[width=0.8\textwidth]{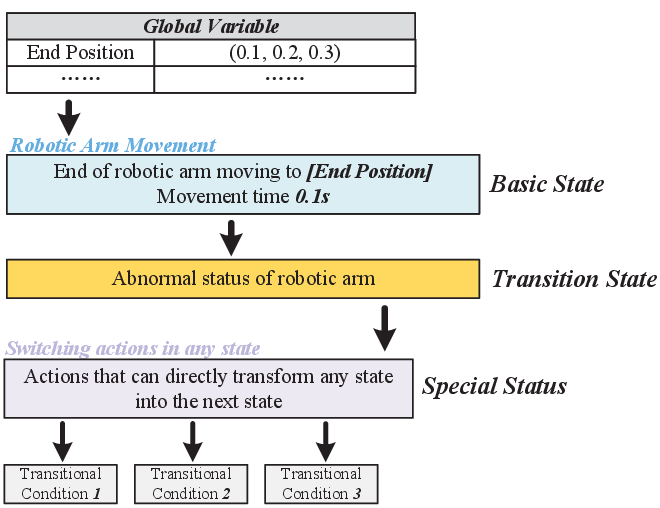}
\caption{The action state machine of the robot.}\label{fig:The action state machine of the robot}
\end{figure}

Using state transition diagrams, we can visualize the robot's action state machine. The diagram consists of state nodes and transition edges, where nodes represent states and edges indicate the relationships between state transitions. The following diagram illustrates in the Fig. \ref{fig:The action state machine of the robot}.

\begin{figure}[H]
\includegraphics[width=\textwidth]{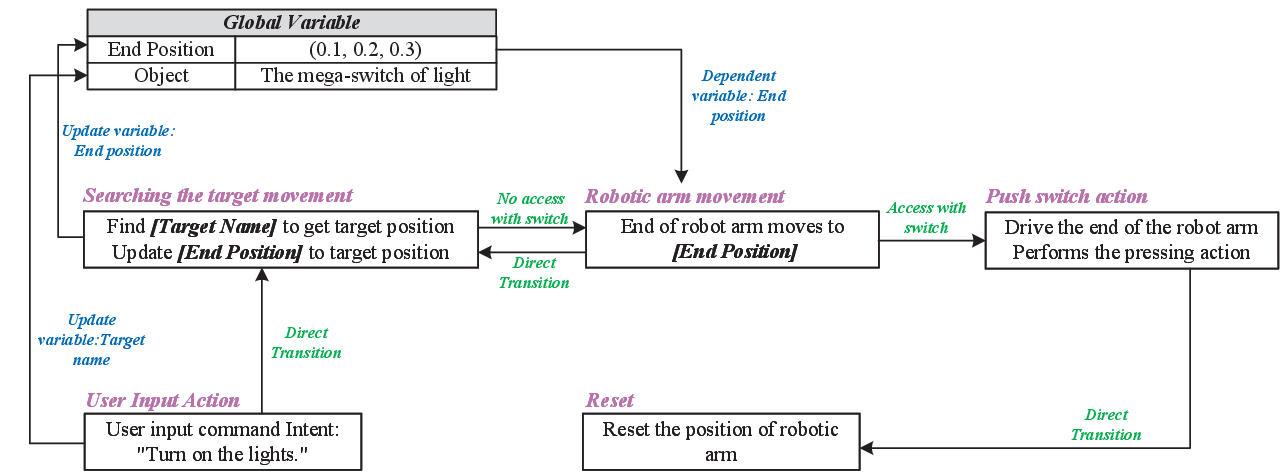}
\caption{The robot arm completes the overall process of turning on the light.}\label{fig:The robot arm completes the overall process of turning on the light}
\end{figure}

The Fig. \ref{fig:The robot arm completes the overall process of turning on the light} depicts the process of the mechanical arm turning on a light. Initially, the user input action is a special state, directly transitioning the action state machine to this state and updating the global variable "target name" to "light." According to the state transition diagram, the next state transitions to the "search for target action" state, updating the search result target location in the global variable "end position." After the "mechanical arm movement action" ends, if the transition condition "not touching the switch" is met, the state continues to transition to the "search for target action" state to continue searching for the "switch" location. If the transition condition "touching the switch" is met, the state transitions to the "press switch action" state. After the "press switch action" state ends, it directly transitions to the "reset action" state to reset the mechanical arm.

In the network communication part, we use the connection-oriented TCP communication protocol, enabling real-time duplex communication between the robot and the remote host. Users can update and control the global variables of the control center's state transition diagram through the network communication link, achieving complete control over the robot.

\subsection{Comprehensive System}
With the integration of a basic control system, a natural language understanding system, a computer vision system, and the establishment of a control communication hub, the robot is now equipped to perform a diverse range of tasks. The control communication hub forms the nucleus for the robot’s flexible control and intelligent interaction capabilities\cite{IoT-BasedImageRecognitionSystemforSmartHome-DeliveredMealServices}.
\subsubsection{Control System Overview}
The foundation of the robot’s functionality lies in its basic control system, which manages direct hardware operations. This system is centered around a 6-degree-of-freedom (6dof) robotic arm, chosen for its efficiency and ability to perform precision tasks typically done by humans. Equipped with torque and position sensors among other devices, the system employs inverse kinematics algorithms to determine the required joint angles, adjusting the arm’s posture via actuators. The real-time monitoring of the arm's state ensures precision and safety in its operations, utilizing ROS (Robot Operating System) as the integration platform for seamless hardware and software interaction.

\subsubsection{Natural Language Understanding System}
This system translates human language into robot-executable instructions. Utilizing the BERT model, text inputs are processed for intent classification and slot filling, converting them into actionable commands. Voice inputs are managed through the OpenAI Whisper model, which transcribes spoken words into text before undergoing intent and instruction translation. This enables the robot to comprehend and respond to human commands in a natural and intelligent manner.

\subsubsection{Computer Vision System}
Essential for environmental perception, this system uses the YOLO algorithm for real-time object detection, such as locating a small ball on a screen. Depth cameras complement this by capturing depth images, which, when merged with RGB data, provide precise three-dimensional object positioning. This integration supplies critical location data to the control system, facilitating accurate object handling\cite{tatikonda2004control}.

\subsubsection{Control and Communication Hub}
Acting as the central processing node, this hub coordinates the data flow between sensors, decision-making algorithms, and actuator controls. It manages TCP connections for real-time updates and command reception from remote sources. This network communication hub not only maintains data synchronization and tracking but also offloads complex computational tasks to a remote host. This reduces the load on embedded systems, allowing them to focus on immediate control and perception tasks, while the remote host handles more complex, non-real-time processing\cite{UnmannedAerialVehicleandArtificialIntelligence}.

\textbf{Architectural Layers:}
The operational architecture is layered to optimize each component’s function:
\\
\textbf{1. Control and Sensor Sampling Abstraction Layer: }Manages basic control of the robotic arm and gathers sensor data.
\\
\textbf{2. Local Application Layer:} Handles real-time object recognition and localization using YOLO.
\\
\textbf{3. Command Extension Layer:} Maintains and operates the control hub.
\\
\textbf{4. Robot Communication Applications Layer:} Facilitates real-time robot status transmission to remote applications and receives commands.
\\
\textbf{5. Remote Application Layer:} Runs the natural language understanding system with less urgency, offering a user control panel for high-level management.

\begin{figure}[H]
    \includegraphics[width=\textwidth]{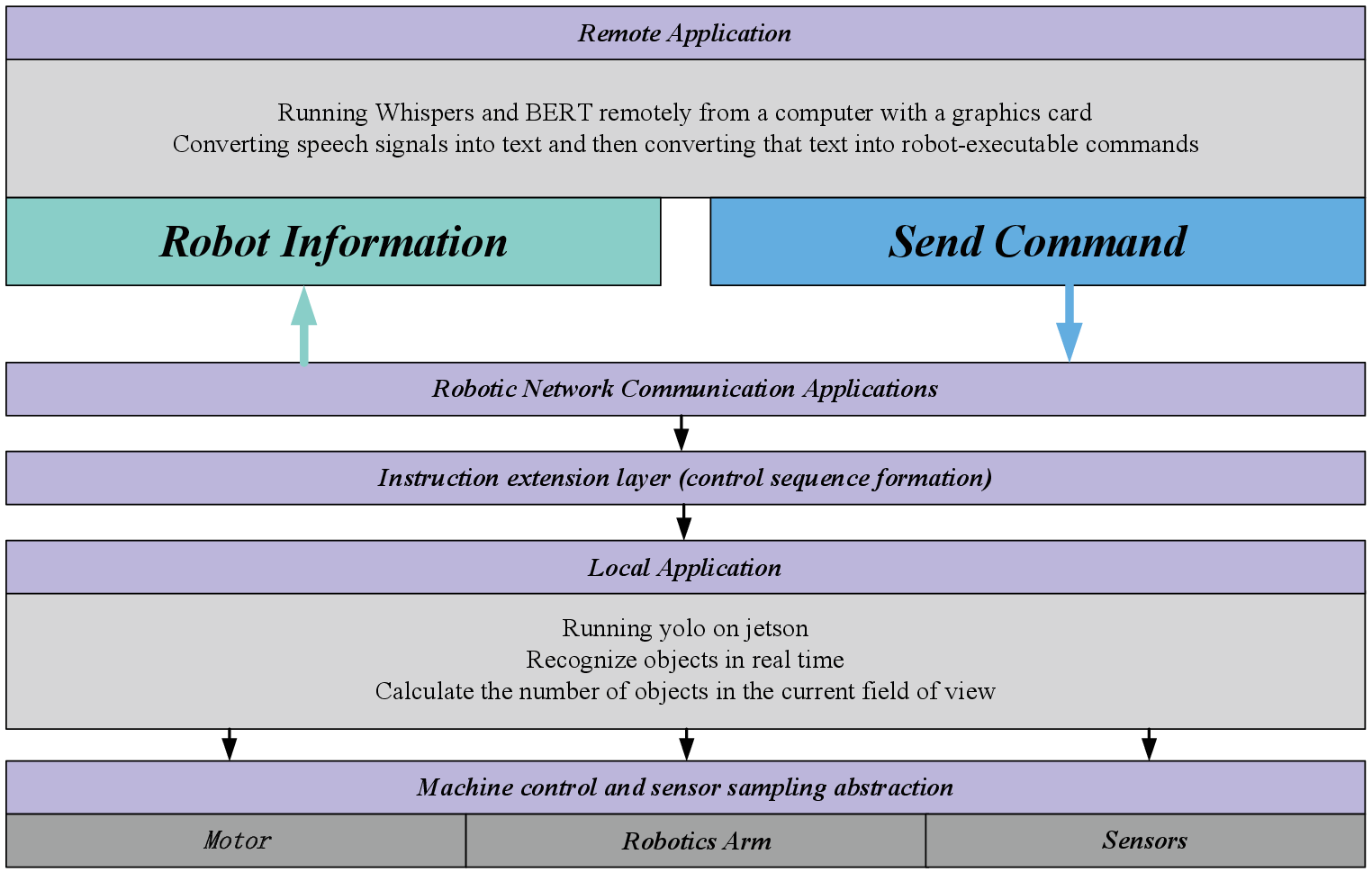}
    \caption{Detailed operational architecture of the control and communication hub.}
    \label{fig:Detailed operational architecture of the control and communication hub}
\end{figure}

\begin{figure}[H]
    \includegraphics[width=\textwidth]{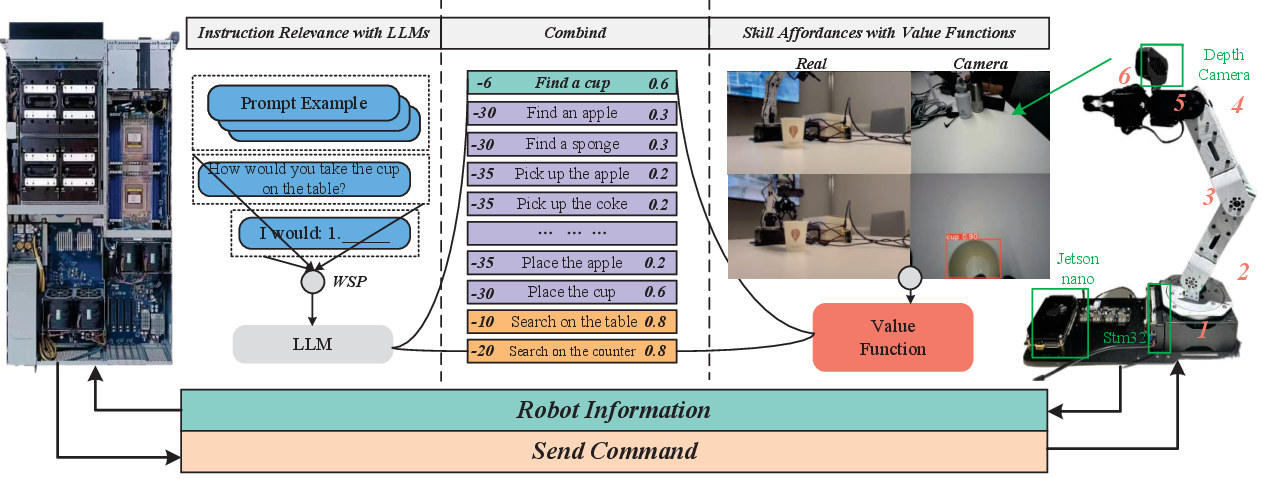}
    \caption{Separation of concerns at the software level.}\label{fig:Separation of concerns at the software level}
\end{figure}

This structured approach ensures that the robot operates efficiently, balancing real-time performance with computational demands to maintain system stability and reliability. The detailed operational architecture is further illustrated in the diagram depicted in Fig.\ref{fig:Detailed operational architecture of the control and communication hub}, and the system’s separation of concerns at the software level is shown in Fig. \ref{fig:Separation of concerns at the software level}, highlighting the division and cooperation across the system’s layers.

\section{Experiments}
\begin{figure}[H]
\includegraphics[width=\textwidth]{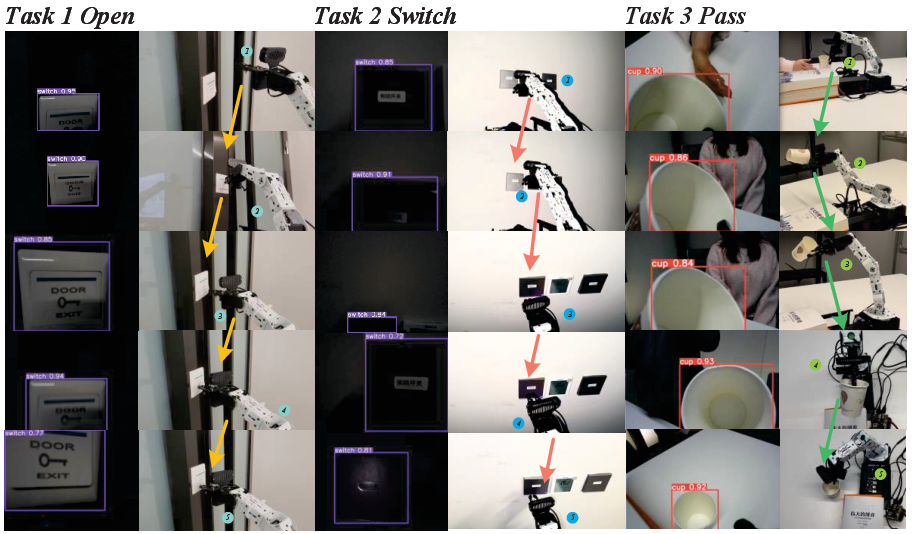}
\caption{three specific tasks.}\label{fig:three specific tasks}
\end{figure}

The robotic arm is tasked with completing three specific tasks in Fig. \ref{fig:three specific tasks}:opening doors, switching lights on and off, and delivering a water cup. The experiment strategy includes controlling the arm via voice commands and capturing the process with a depth camera mounted on the arm, which also provides internal feedback. These tests are structured to progressively evaluate the arm’s performance in increasingly complex tasks.

The first experiment assesses the robotic arm’s ability to integrate multiple technologies by executing simple commands like "open the door" and measuring its completion rate to evaluate feasibility and reliability.

The second experiment expands the tasks to include switching lights on and off. It introduces varied commands and different lighting conditions to test the robotic arm’s adaptability and effectiveness in more complex environments.

The third experiment tests the robotic arm in a living environment, focusing on tasks that require human interaction. This phase evaluates the ease of human-machine interaction, functionality, and the arm's ability to dynamically recognize and respond to changes.

\subsection{Experiment Definition and Design}
\subsubsection{Task 1 Door}
The first experiment tasks the robotic arm with operating office doors, controlled by large rocker switches. Users will issue commands like "open the door," prompting the arm to press the switch to unlock the door. Successful execution is defined by the door being unlocked and ready to be pushed open.
This experiment tests the arm's integration of voice recognition, natural language processing, and robotic control by performing this basic action. Demonstrating the system's functionality with such a task is considered an adequate test of integrated technologies.
The experiment will be conducted 200 times to minimize randomness and ensure reproducibility. Success rates for both command recognition and correct door-opening actions will be recorded.
Successful completion and consistent performance in this initial test will pave the way for subsequent experiments.
\subsubsection{Task 2 Switch}
The second experiment advances from the initial one by broadening the range of control commands and introducing varying lighting conditions. This test involves the robotic arm operating indoor rocker switches, a common fixture. The experiment starts when the tester commands the arm to "turn off the light." Success is determined by the arm's ability to accurately locate and press the far end of the switch and then return to its original position.
Following this, the tester instructs the arm to "turn on the light." In this part of the test, the arm must recognize the switch's position in dim lighting, press the near end of the switch, and revert to its initial state to be considered successful. The room remains devoid of other light sources throughout the test.
Each command, "turn off the light" and "turn on the light," is repeated 200 times to assess the robotic arm's consistency and reliability under altered conditions.
This experiment is designed to be more challenging than the first, testing the arm's adaptability and its proficiency in handling more complex commands and environmental changes.
\subsubsection{Task 3 Cup}
The third and most complex experiment challenges the robotic arm with the task of grasping a water cup and delivering it to the user. This test is designed to evaluate the arm's dynamic recognition capabilities by requiring it to interact with multiple objects and respond to the user's movements. Specifically, the robotic arm must successfully pick up a disposable paper cup from a desk and place it near the user's hand. Successful completion of this task is marked when the cup is accurately delivered to the designated spot.
To add complexity, background noise (B.N.) is introduced to simulate a typical office environment cluttered with items like mice, keyboards, and office supplies. This element tests the robotic arm's ability to perform amidst potential distractions and obstructions. The experiment quantifies success by measuring the clutter area relative to the total area visible to the robot's sensors.
This task will be repeated 200 times under varying conditions of background noise to assess the robotic arm's consistency and adaptability in realistic settings.
This comprehensive test, with its diverse instructions and challenging tasks, serves as a holistic evaluation of the robot's capabilities, pushing its operational limits in dynamic and cluttered environments.
\subsubsection{Compare of Different platform and Versions of YOLO}
In the end of experiment, this paper will also conduct a horizontal comparison of the robotic arm using different versions of YOLO and different core platforms in the first and second experiments.
\subsection{Experiment Details}

\subsubsection{Data of Task 1 Door}
In this experiment, testers used three commands: Command A "Open the door," Command B "Please open the door," and Command C "Please have the door open." Each command was tested 200 times, recording only Correct Speech Recognition (CSR) and Correct Performance (CP). The Execution Rate (ER) was calculated as the sum of correct recognitions and performances divided by the total number of experiments.

\begin{table}[h]
\caption{Record of Task 1}\label{tab2}%
\begin{tabular}{@{}lll@{}}
\toprule
Task 1  & CSR   & CP \\
\midrule
A		& 466			& 428\\
B		& 460			& 421 \\
C		& 463			& 422 \\
ER		& 92.6\%		& 84.3\% \\
\botrule
\end{tabular}
\end{table}

\subsubsection{Data of Task 2 Switch}
In this experiment, the task of turning on the light was conducted under dim lighting conditions (LC), and the task of turning off the light occurred in a bright environment. Testers used four commands each for turning on (A1 "Switch on the light," A2 "Please have the light on," A3 "Turn up the brightness in this room," A4 "It's so dark here, light up please") and off the light (B1 "Switch off the light," B2 "Please have the light off," B3 "Turn down the brightness in this room," B4 "I am gonna sleep, light off please"), recording only Correct Speech Recognition (CSR) and Correct Performance (CP). This method evaluates the robotic arm's ability to respond to diverse commands under varying light conditions, offering insights into its operational efficacy. Experiment results are visually presented in Figure \ref{fig:enter-label}.
\begin{table}[h]
\caption{Record of Task 2}\label{tab3}%
\begin{tabular}{@{}llll@{}}
\toprule
Task 2  & LC   & CSR \\
\midrule
A1		& Dim			& 467          & 383 \\
A2		& Dim			& 461          & 156 \\
A3		& Dim			& 469          & 402 \\
A4		& Dim			& 457          & 392 \\
B1		& Bright		& 472          & 421 \\
B2		& Bright		& 466          & 429 \\
B3		& Bright		& 457          & 414 \\
B4		& Bright		& 452          & 427 \\
\botrule
\end{tabular}
\end{table}

\begin{figure}[h]
    \centering
    \includegraphics[width = 0.98\textwidth]{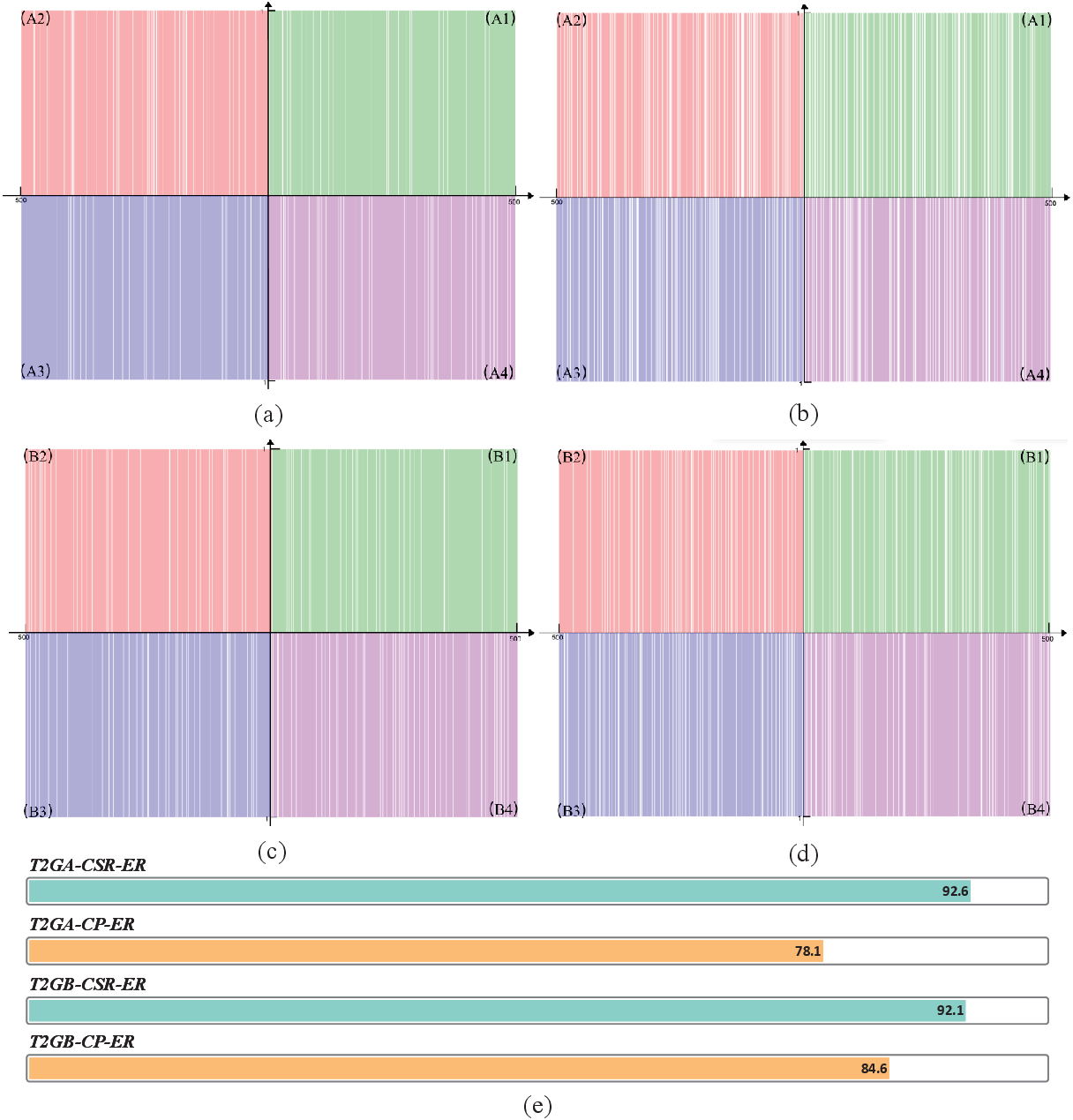}
    \caption{The compare of Speech Recognition and Action Performance in Task 2. The (\textbf{a}) and (\textbf{c}) indicate the CSR-ER. The (\textbf{b}) and (\textbf{d}) indicate the CP-ER.}
    \label{fig:enter-label}
\end{figure}

\subsubsection{Data of Task 3 Cup}
In this experiment, we controlled two main types of variables: command variations and background noise levels, to assess the robot's response. The commands tested were C1 "Please hand me the water cup," C2 "Pass me a cup of water," and C3 "I'm thirsty. I need some water," with only Correct Speech Recognition (CSR) and Correct Performance (CP) being recorded. For the background noise tests, this paper reused data from the command C1 trials in the Orders group. All groups in these tests used command C1, allowing for a focused analysis of how background noise impacts the robot’s command recognition and action execution.

\begin{table}[h]
\caption{Record of Task 3 Group Order}\label{tab4}%
\begin{tabular}{@{}lllll@{}}
\toprule
T3GO    & CSR   & CP    & CSR-ER    & CP-ER \\
\midrule
C1		& 477	& 406    & 95.4\%      & 81.2\%\\
C2		& 469	& 392    & 93.8\%      & 78.4\%\\
C3		& 472	& 386    & 94.4\%      & 77.2\%\\
C4		& 466	& 317    & 93.2\%      & 63.4\%\\
\botrule
\end{tabular}
\end{table}

\begin{figure}[H]
    \centering
    \includegraphics[width = \textwidth]{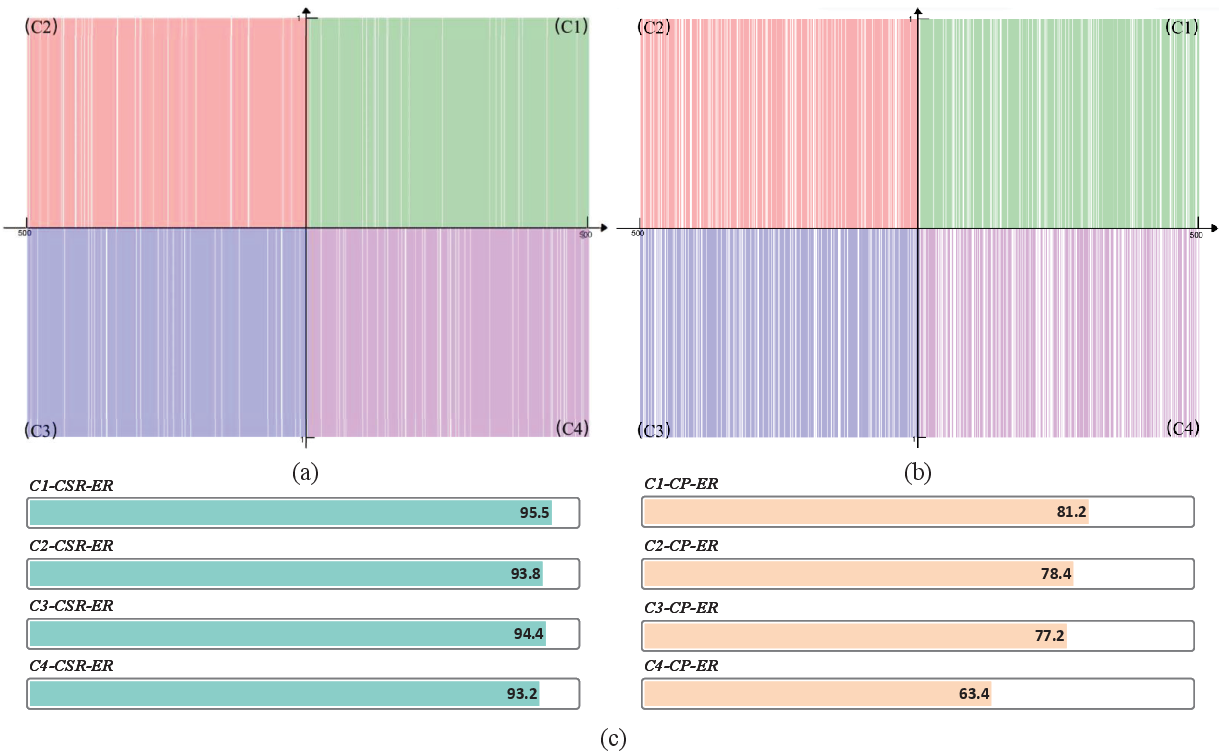}
    \caption{The compare of different commands in Task 3. The (a) indicates the CSR-ER. The (b) indicates the CP-ER.}
    \label{fig:enter-label}
\end{figure}

\begin{table}[h]
\caption{Record of Task 3 Group Background Noise}\label{tab5}%
\begin{tabular}{@{}lllll@{}}
\toprule
B.N.    & CSR   & CP    & CSR-ER    & CP-ER \\
\midrule
0\%     & 476   & 404   & 95.2\%    & 80.8\%\\
25\%    & 465   & 383   & 93.1\%    & 76.2\%\\
50\%    & 464   & 311   & 92.8\%    & 62.2\%\\
75\%    & 467   & 279   & 93.4\%    & 56.0\%\\
\botrule
\end{tabular}
\end{table}

\subsubsection{Data of Compare of Different platform and Versions of YOLO}
This experiment conducts a horizontal comparison to evaluate the effectiveness of the proposed method, focusing primarily on Task 2 and Task 3. Task 2, testing similar objectives as Task 1, was selected because Task 1’s goal was to assess feasibility, which is not revisited in this comparison. The commands in Task 2 included "Please switch on the light" and "Please switch off the light." Task 3 was chosen for its comprehensive testing of the entire solution, using the consistent command "Pass me the paper cup." The robotic arm designed for this study is referred to as "Advanced".

Here are the details of compare:

\begin{table}[h]
\caption{Record of Task 2 Group Algorithm}\label{tab6}%
\begin{tabular}{@{}llllll@{}}
\toprule
T2GA	& LC	& CSR    & CP    &CSR-ER      & CP-ER \\
\midrule
Advanced v7 & Dim & 467 & 383 & 93.4\% & 76.6\%\\
Advanced v7 & Bright & 472 & 421 & 94.4\% & 84.2\%\\
YOLOv5 & Dim & 457 & 327 & 91.4\% & 65.4\%\\
YOLOv5 & Bright & 461 & 406 & 92.2\% & 81.2\%\\
YOLOv7 & Dim & 468 & 322 & 93.6\% & 64.4\%\\
YOLOv7 & Bright & 459 & 396 & 91.8\% & 79.2\%\\
YOLOv8 & Dim & 466 & 386 & 93.2\% & 77.2\%\\
YOLOv8 & Bright & 469 & 431 & 93.8\% & 86.2\%\\
\botrule
\end{tabular}
\end{table}

\begin{table}[h]
\caption{Record of Task 2 Group Platform}\label{tab7}%
\begin{tabular}{@{}llllll@{}}
\toprule
T2GP    & LC   & CSR    & CP    & CSR-ER    & CP-ER \\
\midrule
Nano B01 & Dim & 467 & 383 & 93.4\% & 76.6\%\\
Nano B01 & Bright & 472 & 421 & 94.4\% & 84.2\%\\
Orin NX & Dim & 457 & 327 & 91.4\% & 65.4\%\\
Orin NX & Bright & 461 & 406 & 92.2\% & 81.2\%\\
Orin AGX & Dim & 468 & 322 & 93.6\% & 64.4\%\\
Orin AGX & Bright & 459 & 396 & 91.8\% & 79.2\%\\
\botrule
\end{tabular}
\end{table}

\begin{table}[h]
\caption{Record of Task 3 Group Algorithm}\label{tab8}%
\begin{tabular}{@{}lllll@{}}
\toprule
T3GA    & CSR   & CP    & CSR-ER    & CP-ER \\
\midrule
Advanced & 472 & 421 & 94.4\% & 84.2\%\\
YOLOv5 & 461 & 404 & 92.2\% & 80.8\%\\
YOLOv7 & 469 & 396 & 93.8\% & 79.2\%\\
YOLOv8 & 476 & 417 & 95.2\% & 86.2\%\\
\botrule
\end{tabular}
\end{table}

\begin{table}[h]
\caption{Record of Task 3 Group Platform}\label{tab9}%
\begin{tabular}{@{}lllll@{}}
\toprule
T3GP    & CSR   & CP    & CSR-ER    & CP-ER \\
\midrule
Advanced & 477 & 406 & 95.4\% & 81.2\%\\
Orin AGX & 473 & 399 & 93.2\% & 79.8\%\\
Orin NX & 479 & 411 & 95.8\% & 82.2\%\\
\botrule
\end{tabular}
\end{table}

\subsection{Discussion of the Experiment's Results}
\subsubsection{Discussion of Task 1}
The robotic arm, based on the method proposed in this paper, demonstrated commendable performance in executing tasks under the condition of receiving different commands. It exhibited high execution rates in both speech recognition and performance execution, which were 93.1\% and 84.5\%, respectively. These data support and validate the feasibility of the proposed solution, and the functional stability has also been confirmed through repeated experimental testing.

\subsubsection{Discussion of Task 2}
Experimental data from the task of switching lights on and off revealed that the success rate for turning off the light was over 5\% higher than for turning it on. This discrepancy is attributed to indoor lighting conditions, which significantly impacted the visual recognition system's effectiveness. Despite varying the commands—from slight alterations to those not directly mentioning the light status—the language recognition and action execution success rates remained consistent. This confirms the robotic arm’s robust ability to understand simple natural language commands.

\subsubsection{Discussion of Task 3}
These findings reveal that while language recognition rates remain high and stable across various commands, there is a more variable execution rate in performing actions, particularly with a notable decrease for command C4. This suggests that although the robot consistently understands the commands, its ability to execute specific actions may be affected by the command's precision or complexity. The experiment, divided into two main groups, showed consistent speech recognition rates across all subgroups. However, performance execution rates were similar among groups C1, C2, and C3, but significantly lower for group C4, which used less precise commands involving the term "water" without specifying "paper cup," leading to potential ambiguity. Despite this, a 63.4\% execution rate was achieved, helped by the absence of any background noise and only the water cup being present, prompting the robot to act on the vague command. For future improvements, enhancing the robot’s inferential capabilities in its natural language model is recommended.
The impact of background noise on the robot's performance was particularly significant in affecting action execution. While the robot's language recognition was relatively unaffected by increasing noise levels, the action execution success rate decreased markedly with higher noise, indicating a greater sensitivity of the robot’s operational capabilities to auditory disturbances.

\subsubsection{Discussion of the Compare with Different Versions and Platform}
This comparison across different platforms reveals that despite specification upgrades from the Nano to the AGX and NX versions of Orin, there are no significant improvements in speech recognition or performance execution, with fluctuations under 5\%. Therefore, the more cost-effective and readily available Nano platform is preferred due to its comparable usability and performance.
Regarding YOLO versions, the YOLO models are run locally on the Nano B01, necessitating a balance between model size and computational demands. Advanced YOLOv7 shows excellent performance in the experimental tests, achieving execution rates of 76.6\% and 84.2\%.
The comprehensive testing across three tasks confirms the feasibility and stability of the proposed solution integrating ASR, NLP, and CV technologies. It demonstrates reliable robustness, stability, and transferability, meeting the operational demands of edge devices.

\section{Conclusion}
This paper proposes a framework for a ROS-driven robot, utilizing the Jetson Nano B01 as its core platform, equipped with a depth camera, and integrating technologies such as Whisper, BERT, and YOLO. Moreover, through three tasks of opening doors, switching lights, and passing paper cups, the multi-task execution capabilities of this desktop-level robot were thoroughly validated from different perspectives and commands. While there are still many issues to address compared to the initial inspiration—the gizmo from the "Rick and Morty" animation—such as further miniaturization, future addition of speakers for human-machine dialogue, and alleviating the computational strain on edge devices to deploy newer versions of models, the conclusion that can be drawn at this point is that the feasibility of an integrated solution for the fundamental technologies required by the initially conceived desktop-level robot with multimodal input has been validated. This also demonstrates the potential of desktop and edge devices to handle complex multi-tasks for such robots.

\section*{Declarations}

\begin{itemize}
\item Funding
National College Students' Innovation and Entrepreneurship Training Program No.202210022067

\item Competing interests 
No potential and competing conflict of interest was reported by the authors.

\item Ethical and informed consent for data used
Not applicable.

\item Consent for publication
Not applicable.

\item Data availability and access
The authors confirm that the data supporting the findings of this study are available within the article

\item Author contribution statement
Conceptualization: Wenbin Pan, Haohua Que; 
Methodology: Haohua que, Wenbin Pan, Hao Luo, Jie Xu, Pei Wang, Li Zhang; 
Formal analysis and investigation: Haohua Que, Wenbin Pan, Jie Xu; 
Writing - original draft preparation: Haohua Que Wenbin Pan; 
Writing - review and editing: Pei Wang, Hao Luo, Jie Xu, Wenbin Pan; 
Funding acquisition: Haohua Que, Pei Wang;
Supervision: Pei Wang, Li Zhang.
\end{itemize}






\end{document}